\documentclass[12pt]{spieman}  
\usepackage{amsmath,amsfonts,amssymb}
\usepackage{graphicx}
\usepackage{setspace}
\usepackage{tocloft}
\usepackage{multirow}
\usepackage{lineno}

\title{Efficient human-in-loop deep learning model training with iterative refinement and statistical result validation}

\author[a]{Manuel Zahn}
\author[a,b,*]{Douglas P. Perrin, Ph.D.}
\affil[a]{Harvard School of Engineering and Applied Sciences, Cambridge, MA 02138 USA}
\affil[b]{Department of Cardiac Surgery, Children’s Hospital Boston, Harvard Medical School, Boston, MA 02115 USA}

\cftpagenumbersoff{figure}
\cftpagenumbersoff{table} 
\begin{document} 
\maketitle

\begin{abstract}\\[10pt]
Annotation and labeling of images are some of the biggest challenges in applying deep learning to medical data. Current processes are time and cost-intensive and, therefore, a limiting factor for the wide adoption of the technology. Additionally validating that measured performance improvements are significant is important to select the best model. In this paper, we demonstrate a method for creating segmentations, a necessary part of a data cleaning for ultrasound imaging machine learning pipelines. We propose a four-step method to leverage automatically generated training data and fast human visual checks to improve model accuracy while keeping the time/effort and cost low. We also showcase running experiments multiple times to allow the usage of statistical analysis. Poor quality automated ground truth data and quick visual inspections efficiently train an initial base model, which is refined using a small set of more expensive human-generated ground truth data. The method is demonstrated on a cardiac ultrasound segmentation task, removing background data, including static PHI. Significance is shown by running the experiments multiple times and using the student’s t-test on the performance distributions. The initial segmentation accuracy of a simple thresholding algorithm of 92\% was improved to 98\%. The performance of models trained on complicated algorithms can be matched or beaten by pre-training with the poorer performing algorithms and a small quantity of high-quality data. The introduction of statistic significance analysis for deep learning models helps to validate the performance improvements measured. The method offers a cost-effective and fast approach to achieving high-accuracy models while minimizing the cost and effort of acquiring high-quality training data.
\end{abstract}


{\noindent \footnotesize\textbf{*}Douglas P. Perrin,  \linkable{douglas.perrin@childrens.harvard.edu} }

\begin{spacing}{2}   
\section{Introduction}
\label{sect:intro}  
In recent years, deep learning technology has gained a lot of popularity. While showing remarkable results, the development process can still be costly and slow. Therefore, research has been focused on increasing the speed of model training, decreasing the deep learning model sizes, and reducing the amount of training data required. An early approach was presented by Han et al. in 2016. They reduced the storage required by AlexNet\cite{NIPS2012_c399862d} by 35x and the size of VGG-16\cite{7486599} by 49x, without loss of accuracy by using network pruning\cite{han2016deep}. Frankle and Carbin presented their “Lottery Ticket Hypothesis” to identify subnets that learn faster than the original network and reach higher test accuracies\cite{Frankle2018}. Choi et al. introduced the concept of "data echoing," which decreased in wall-clock time for ResNet-50\cite{he2016deep} on ImageNet\cite{deng2009imagenet} by a factor of 3.25 when reading training data over a network\cite{Choi2019}.

Besides model training, data labeling is one of the most time-consuming and expensive tasks when developing supervised deep learning models. For many medical applications, models are also required to achieve very high accuracy levels. Experts often spend hours annotating images. Additionally large number of images is need to be labeled before even an initial baseline model can be trained. Currently, there is also no definition of what encompasses a baseline medical imaging dataset in terms of adequate data volume and image quality for medical machine learning problems\cite{Harvey2019}. These labeling challenges limit a wider application of deep learning technology and make the process of model development long and expensive.


In this paper, we focus on creating a process to reduce the amount of expensive training data needed to build segmentation masks. This is showcased for single cone ultrasound images, where it can be leveraged as a removal tool for burned-in patient health data. We present a four-step method to drastically reduce the amount of high-quality human labels necessary to create accurate deep learning segmentation models. This is done by leveraging poor quality automatically generated labels and fast human visual sorting. The process shortens the time for model development and enables use cases where the generation of high-quality labels is difficult and expensive to obtain. We also showcase that simple algorithms can be improved to match the performance of more complicated algorithms, thus reducing the necessity to spend a lot of development time on complicated initial segmentation algorithms. The idea of reducing the necessity for large amounts of expensive training data has previously been addressed by using pre-trained model weights. Many applications leverage weights trained on large datasets like ImageNet\cite{Russakovsky2015} and apply them to other problems in a similar image domain. This method is not justifiable when working with healthcare data since most large-scale image datasets don't contain medical image data. For example, existing patterns of clear edges and textures and color space gradients that are present in most natural images are lacking in many medical image types. The impact of this significant domain difference on the usage of pre-trained weights of large-scale natural image datasets for medical image purposes is not well characterized.

In our work, we will therefore train our base models from scratch and then refine them with further training on a small hand segmented ground-truth dataset. R. Barth et al. presented a similar concept for agricultural segmentation. In that work, a deep learning model was trained on a synthetic bell pepper image dataset and then was refined for better performance with additional training on a small hand segmented real-world image dataset\cite{Barth2019}. In contrast to R. Barth et al., the time-intensive creation of synthetic images will be replaced by exploiting computer vision algorithms of varying complexity and a fast manual sorting to create a training set for a base model. In total, this method enables a quick and inexpensive (in both human and computational costs) way of creating deep learning training data.

The described approach in this paper can be used as a precursor to other Machine Learning approaches. The masking of medial image data also helps to remove fiducials and other background objects. Those pose the risk of creating prediction biases which can have catastrophic outcomes for medical deep learning models if deployed in a clinical setting. Data preparation and preprocessing are, therefore, among the most critical tasks to ensure unbiased training data. Often it is not possible to receive raw image data without features that would indicate the image source. Examples are manufacturer logos, product names, specific locations of saturation bars, visual trackers, and PHI data like patient name or age that are overlaying the image data. When training deep learning models on datasets containing multiple data sources, visual indicators in the data like this can cause strong model biases and therefore need to be addressed. 

\begin{figure}
\begin{center}
\begin{tabular}{c}
\includegraphics[width=0.95\textwidth]{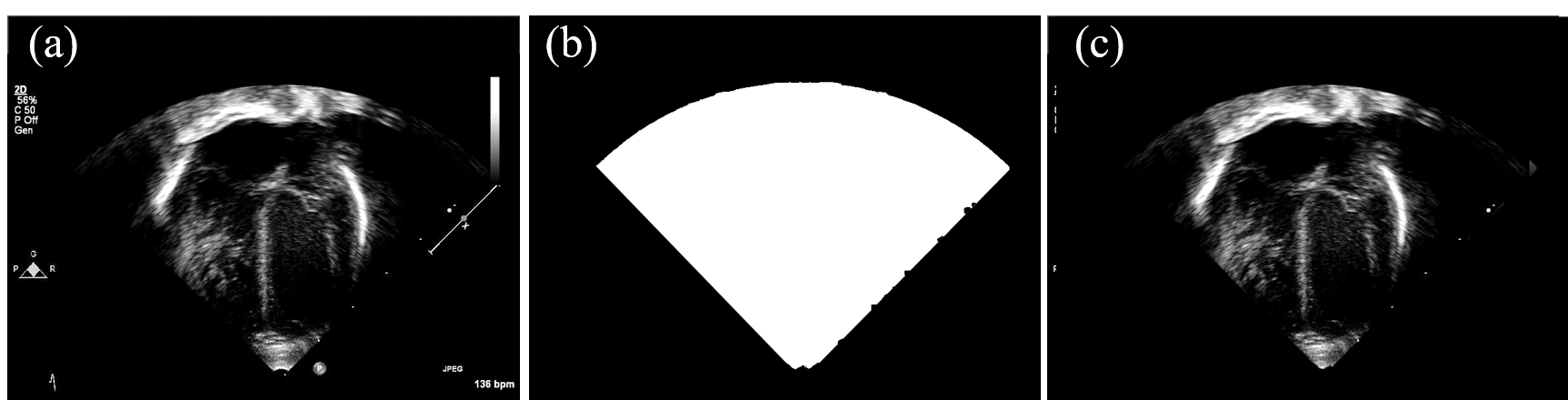}
\end{tabular}
\end{center}
\caption 
{ \label{fig:fig_1}
(a): The input image containing non-echo clutter, including potential PHI information, (b): A binary mask, only containing ultrasound image data, (c): The input image without any potential bias and PHI, after the mask was applied. } 
\end{figure}

In this work, the cardiac ultrasound background removal task will be used to showcase our method. The presented concepts can also be applied to fetal and other single cone ultrasound image data. In clinical ultrasound images, static and dynamic elements like patient information, saturation bars, EKG, or patient information are combined in one frame with the ultrasound data (Fig.~\ref{fig:fig_1}, a). Due to this bias, cone segmentation is a necessary preprocessing step for deep learning models on ultrasound data. To do this, we create and apply a binary mask (Fig.~\ref{fig:fig_1}, b) to the input image (Fig.~\ref{fig:fig_1}, c). The training sets for the base models will be generated using computer vision algorithms of varying complexity, to compare how their performance improves using the presented method. Expensive hand segmented data is only used for the model refinement step.

\section{Methods}
\subsection{Data Collection}

\begin{figure}
\begin{center}
\begin{tabular}{c}
\includegraphics[width=0.95\textwidth]{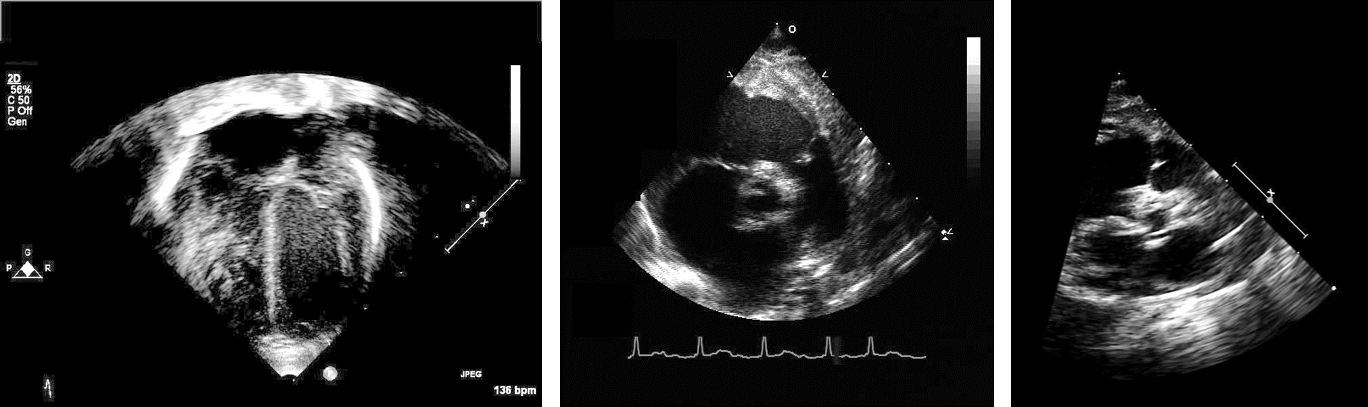}
\end{tabular}
\end{center}
\caption 
{ \label{fig:fig_2}
A few example frames of different heartbeat sequences, showing the variation in rotation, scale, and structure included in the dataset.} 
\end{figure}

The dataset for this paper came from Children’s Hospital Boston and the Pediatric Heart Network. It consists of 1,666 cardiac sequences collected from children’s hospitals across the US. Each sequence consists of 10 - 120 frames and shows at least one heartbeat. The frames are provided in the Portable Network Graphics (PNG) format with no segmentation or labeling. Within the sequences, the ultrasound cones vary in scale, location, rotation, and saturation. Some cones are occluded or cut off, and in some sequences, background objects (static and dynamic) overlay the cone structure. The image sequences come in 7 different sizes and vary between a maximum of 800 x 600 pixels and a minimum of 480 x 430 pixels. A few selected images are in Fig.~\ref{fig:fig_2}.

\subsection{The Model Refinement Method}

\begin{figure}
\begin{center}
\begin{tabular}{c}
\includegraphics[width=0.95\textwidth]{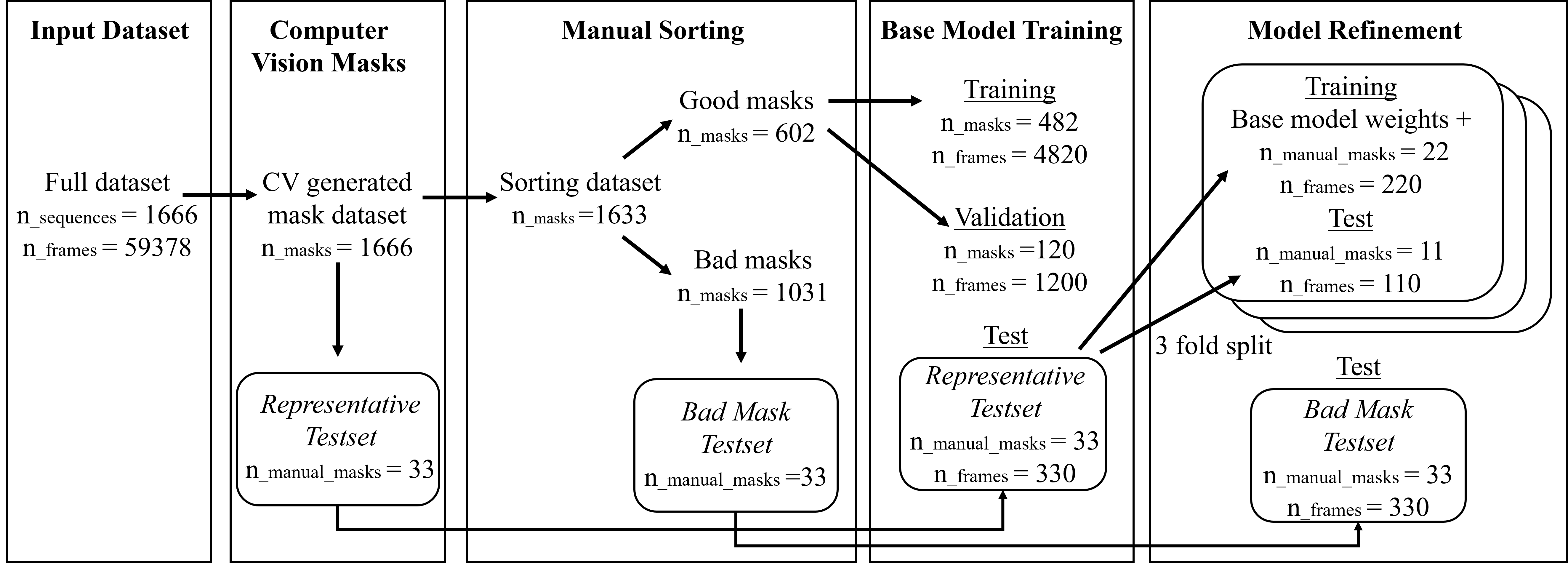}
\end{tabular}
\end{center}
\caption 
{ \label{fig:fig_3} The full model refinement method.} 
\end{figure}

Our model refinement method consists of four steps. In the first step, we create initial training masks using computer vision to segment the ultrasound cones. We hypothesize that the output of those masking algorithms can be used to train mediocre-performing base models, which can be refined to a better accuracy using a small set of expensive hand-segmented data. In the process, the variance between simpler and more complex initial computer vision segmentation algorithms is reduced. The advantage of this refinement approach is that it saves effort on computer vision algorithm development, ground truth generation, or both. In this work, the performance of a simple thresholding, a filled thresholding and a convex hull algorithm will be used to train three base models. A test set of 33 images is randomly selected out of the dataset. The images are hand segmented to create a ground truth test set, which is representative of the entire dataset (Labeled: Representative Testset). This set will be used to evaluate the model accuracies, and a subset will be used to perform the refinement step. The second step is a fast manual quality sorting of the label masks. Manual sorting utilizes the efficiency of the human visual system to distinguish between good and bad segmentation results. Simple in/out sorting by hand is much faster than hand segmentation. The images classified as good in the visual check are then used as the training set for the base models. From images classified as bad during manual sorting, another test set of 33 different random images is selected. These images are hand segmented to create a test set of only bad images (Labeled: Bad Mask Testset). This set allows model evaluation on cases where the automated segmentation of the computer vision algorithms failed. In the third step, the masks classified as good are used to train base models. The performance of those models on the test sets allows a comparison of the model improvements using our proposed method. In the final fourth step, a subset of the hand segmented "Representative Testset" images is used to refine the base models. The final model performance evaluations come from evaluating them on the "Bad Mask Testset" set. Figure~\ref{fig:fig_3} shows the entire refining method. We run the modeling experiments six times and propose to leverage statistical analysis on the resulting model performance distributions to receive measures of experiment significance.

\subsection{Results}
\subsubsection{Dataset creation using computer vision algorithms}

\begin{figure}
\begin{center}
\begin{tabular}{c}
\includegraphics[width=0.95\textwidth]{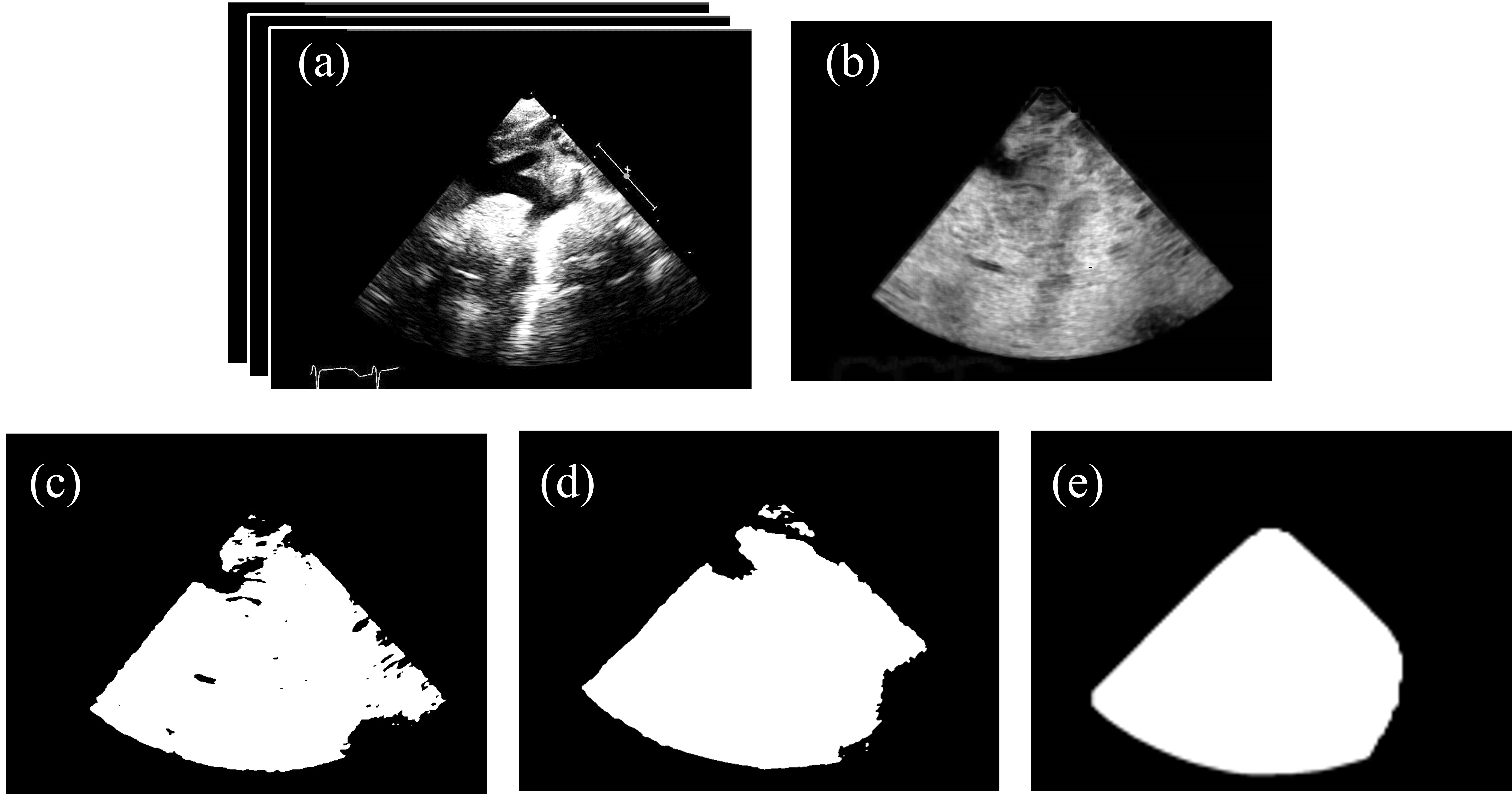}
\end{tabular}
\end{center}
\caption 
{\label{fig:fig_4}
The pipeline to create the training masks, (a): n raw cardiac frames, (b): the mean of n-1 difference images, (c): adaptive thresholding creating a binary mask, (d) the masks after applying a flood fill algorithm, (e) the mask after applying a convex hull algorithm}
\end{figure}

We used different computer vision algorithms to create three initial unsorted training datasets. To remove static background objects, we leverage the sequential nature of the given images. By subtracting subsequent frames, only image parts containing changes are kept. N frames of a sequence (Fig.~\ref{fig:fig_4}, a) therefore generate N-1 difference images. Those can then be combined into a single image by calculating the mean (Fig.~\ref{fig:fig_4}, b) over the N-1 resulting difference images. This mean difference image is then used as a base to sequentially create the three masking types used for the performance comparison in this paper. The simplest segmentation mask is obtained by applying an adaptive threshold on the base image (Fig.~\ref{fig:fig_4}, c). Since it was created with the combined mean image, the mask is valid for the whole image sequence. In a next step, a flood fill algorithm is applied to the thresholding mask. This results in a mask, where holes in the structure are closed, and the more accurate ”filled threshold” mask (Fig.~\ref{fig:fig_4}, d) is created. In the final step a convex hull algorithm is applied, achieving an even better result (Fig.~\ref{fig:fig_4}, e). These steps are performed on the entire dataset, creating a set of segmentation masks for base model training. This is relatively low-cost since a hand segmentation or creation of synthetic ultrasound images is not necessary. By applying the same computer vision algorithm to the "Representative Testset," an initial performance of 91.87\% for the thresholding masks, 95.74\% for the filled thresholding masks and 97.24\% for the hull masks can be calculated. The accuracy is defined as the sum of all True Positive (TP) and True Negative (TN) pixels divided by the number of pixels in the image.

\subsubsection{Manual sorting of images}
Sorting leverages the human visual assessment by categorizing all masks into two categories. Masks that cover the majority of the ultrasound cone are considered good and added to the training set. Masks that include parts of the background or the masked region that miss a lot of the ultrasound cone are regarded as failures of the automated segmentation and are removed. The manual assessment proved to be much faster than the creation of hand-segmented ground truth masks or synthetic ultrasound data while still leveraging human decision capabilities. Hand segmenting a single image takes on average three minutes and is, therefore, more time-intensive. For the whole 1666 images, 4998 minutes or 83.3 hours of manual segmentation would be required. By comparison, the in/out sorting for the masks created by the computer vision algorithm on the full dataset was performed in less than 20 minutes. After sorting, 602 masks of the sequences were classified as good and kept as the training set for the base models, which is only 37\% of the initial dataset. Therefore 63\% of the segmentations created using traditional computer vision algorithms failed the visual assessment.

\subsubsection{Training of base models}
\begin{figure}[htb]
\begin{center}
\begin{tabular}{c}
\includegraphics[width=0.95\textwidth]{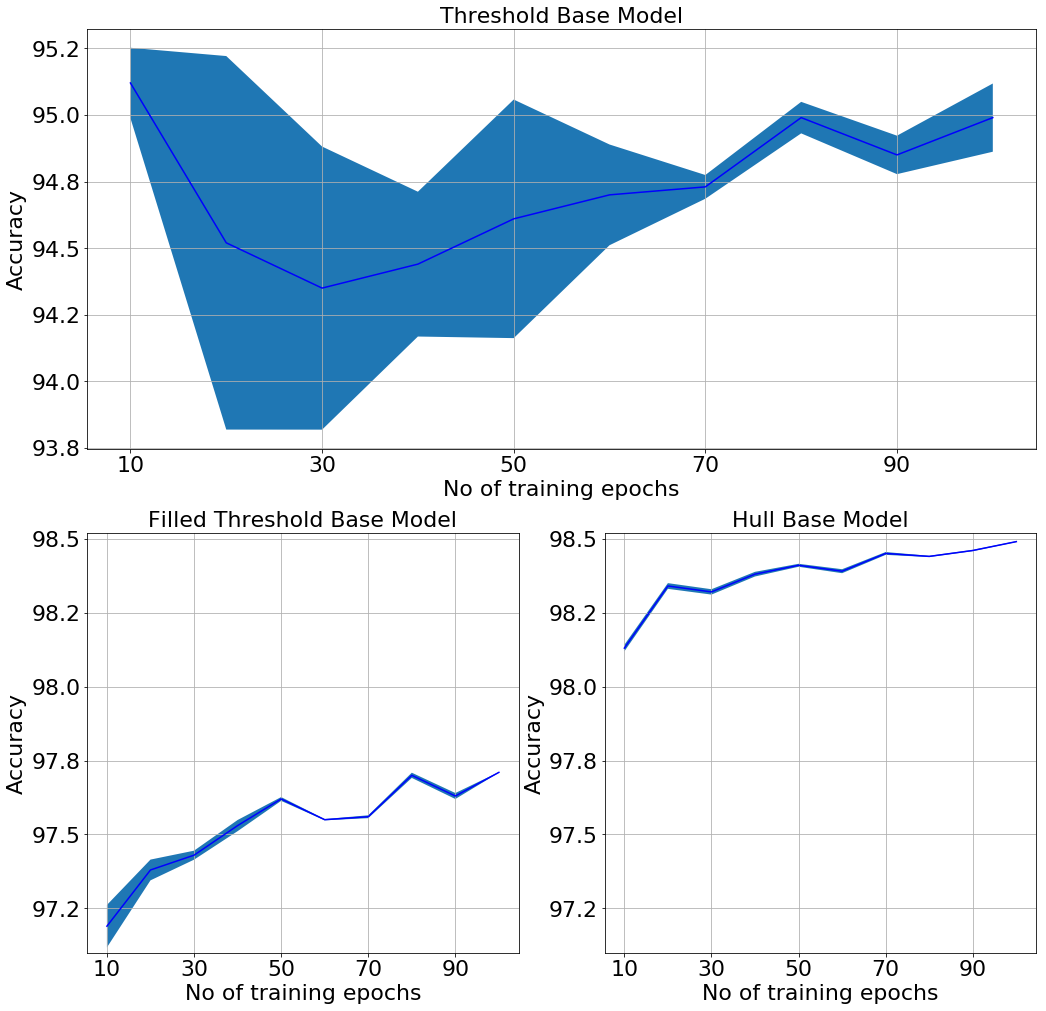}
\end{tabular}
\end{center}
\caption 
{ \label{fig:fig_5}
Mean training accuracy and variance of the base models every ten epochs evaluated on the "Representative Testset". Upper row: the threshold model, lower row left: the filled threshold model, lower row right: the hull model}
\end{figure}

Training of the base models was performed using the set without bad masks. We choose the U-Net structure\cite{ronneberger2015} as a model architecture. It is a convolutional network designed for biomedical image segmentation. It consists of a contracting path and an expansive path, resulting in a u-shaped structure. Our experiments ran on the Harvard FAS Research Computing cluster using 4 GPUs with a 32GB memory pool for all cores. The training uses a mean squared error Adam optimizer\cite{Kingma2015} with a learning rate of 1e-5 and a batch size of 8 images. We reserved twenty percent of the input data for validation, and the image sizes are 256x256 pixels. Since there is only one mask per sequence, multiple images of each sequence could be used to train with the same mask. We chose ten random frames of each sequence, each of which used its corresponding mask for the model training. The learning curve is evaluated on the hand segmented "Representative Testset" every 10th epoch step. The experiment ran six times, choosing a different set of ten random frames per sequence for each run. This allows to apply statistical significance calculations by comparing the model performance distributions of the runs. The curves show the mean accuracy and the variance over the six runs for the three input mask types. Figure~\ref{fig:fig_5} shows the training accuracy and variance of the base models evaluated on the test set. The variance of the six model runs is high in the beginning but gets smaller after more training epochs. The accuracy reaches an average of 94.99\% for the thresholding, 97.71\% for the filled thresholding, and 98.49\% for the hull masks after 100 epochs. This is an increase of 3.12\%, 1.97\%, and 1.25\% respectively compared to the computer vision-based algorithms. This also shows that there is still a large variance with a performance difference of 3.5\% between the simplest computer vision thresholding input masks and the more sophisticated convex hull input masks. We hypothesized that the performance could be further improved whilst also reducing this input mask-dependent performance variance.

\subsubsection{Model refinement with hand segmented dataset}
\begin{figure}[htb]
\begin{center}
\begin{tabular}{c}
\includegraphics[width=0.95\textwidth]{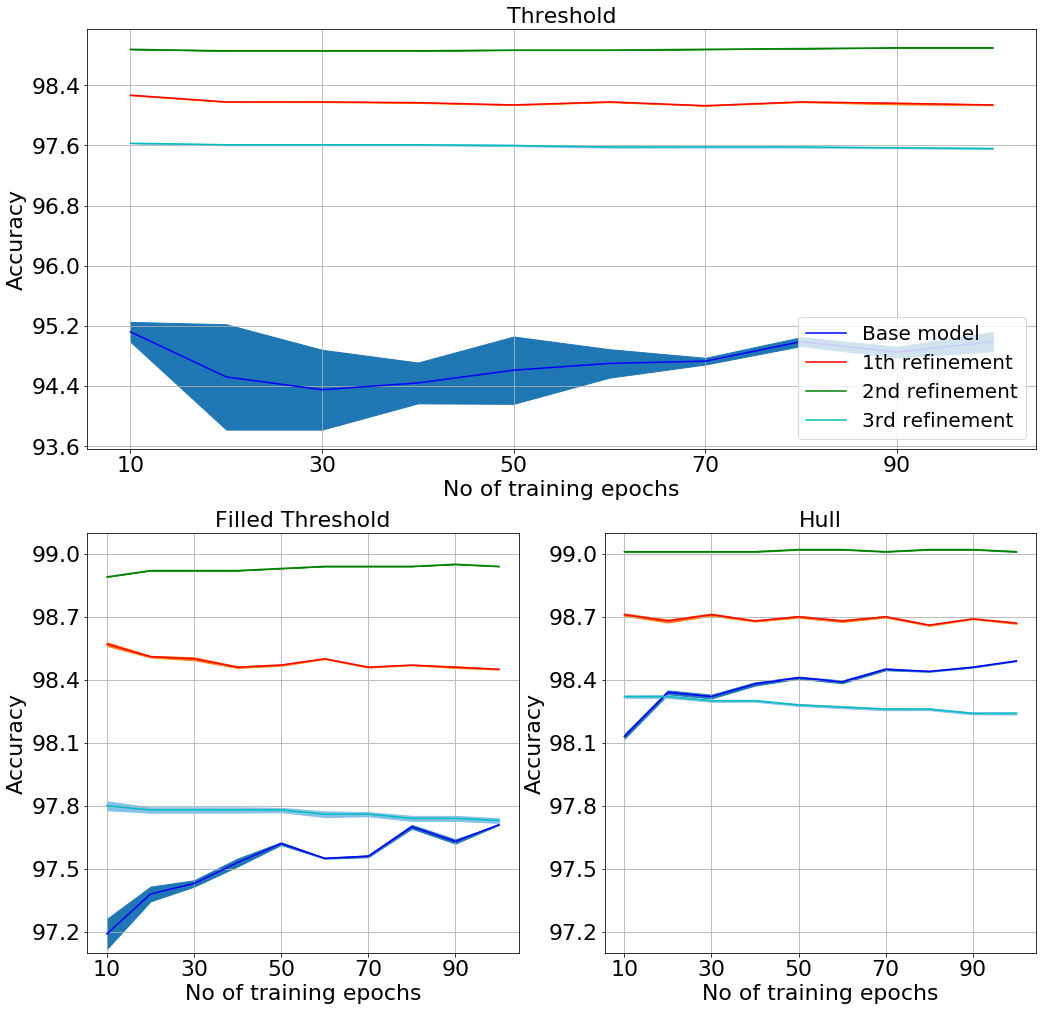}
\end{tabular}
\end{center}
\caption 
{ \label{fig:fig_6}
Mean training accuracy and variance of the base models and the three refinement splits every ten epochs evaluated on the "Representative Testset". Upper row: the threshold model, lower row left: the filled threshold model, lower row right: the hull model}
\end{figure}

The performance of the base models already showed some improvement over the computer vision algorithms. Our hypothesis for the model refinement was that a small accurate hand segmented dataset could be used to improve the performance even further. The validation should be shown by performing a stastistical analysis on the performance distribution of the multiple training runs. The concept would enable simpler computer vision algorithms to achieve accuracies on the level of more complex algorithms or more significant amounts of expensive training data. The idea is to use a part of a test set to refine the base models while evaluating the new model with the remaining test set. Therefore the "Representative Testset" is divided into three parts; two-thirds refine the base models through additional training while one-third provide an evaluation of the performance. This is repeated for all three possible combinations (1st fold, 2nd fold, and 3rd fold). Every experiment is run six times trained on different random frames, creating a mean and variance over the training epochs. The results can be seen in Fig.~\ref{fig:fig_6}. Since ten frames come from each sequence, each training set contains 220 training cases, with 20\% used for validation and 110 test cases to evaluate the performance. This also allows comparing two models by calculating the p-value, as described by the student t-test\cite{Student1908}, using the six runs of each model. The mean accuracies and 95\% confidence intervals of the base model and three refinements can be seen in Table~\ref{tab:table_1}. When evaluating the statistical significance of the performance between two models, a Holm-Bonferroni correction is applied to control for the family-wise error rate. All improvements over the base model are significant if the sorted p-values are smaller than the respective $\alpha_{Holm-Bonferroni}$ value. The $\alpha_{Holm-Bonferroni}$ comparison values for those experiments are 8.3e-03, 1.00e-02, 1.25e-02, 1.67e-02, 2.50e-02 and 5.00e-02 respectively and comparisons that fulfill the statistical significance criteria are marked with an asterisk. The results can be seen in (Table~\ref{tab:table_2}). The average accuracy of the refined models outperforms the base models for all cases, besides the third refinement of the hull mask. And all measured changes, besides the third refinement of the filled threshold, are statistically significant with p-values below the respective $\alpha_{Holm-Bonferroni}$. Overall it can be seen that the difference between the initial computer vision masking algorithms is largely reduced when comparing the refinement models. The initial accuracy difference of 3.5\% between the threshold and hull base models is reduced to below 1\% for all folds and only 0.15\% for the second refinement. This shows that the refinement method enables using simpler computer vision algorithms with less time of development to achieve similar performance accuracies as more complex methods.

\begin{table}[htb]
\caption{Average model accuracies (n=6) and 95\% confidence intervals}
\label{tab:table_1}
\begin{center}   
\begin{tabular}{|c|cc|cc|cc|}
\hline
\multicolumn{1}{|c|}{} & \multicolumn{2}{c|}{Threshold} & \multicolumn{2}{c|}{Filled Threshold} & \multicolumn{2}{c|}{Hull} \\ \cline{2-7} 
\multicolumn{1}{|c|}{Model} & \multicolumn{1}{c|}{mean acc} & \multicolumn{1}{c|}{95\% confid.} & \multicolumn{1}{c|}{mean acc} & \multicolumn{1}{c|}{95\% confid.} & \multicolumn{1}{c|}{mean acc} & \multicolumn{1}{c|}{95\% confid.} \\ \hline
Base & \multicolumn{1}{c|}{94.99} & (94.58, 95.41) & \multicolumn{1}{c|}{97.71} & (97.68, 97.75) & \multicolumn{1}{c|}{98.49} & (98.44, 98.54) \\ \hline
1st ref.& \multicolumn{1}{c|}{98.14} & (98.09, 98.2) & \multicolumn{1}{c|}{98.47} & (98.41, 98.54) & \multicolumn{1}{c|}{98.7} & (98.62, 98.78) \\ \hline
2nd ref.& \multicolumn{1}{c|}{98.87} & (98.84, 98.9) & \multicolumn{1}{c|}{98.93} & (98.92, 98.95) & \multicolumn{1}{c|}{99.02} & (98.97, 99.06) \\ \hline
3rd ref.& \multicolumn{1}{c|}{97.6} & (97.49, 97.71) & \multicolumn{1}{c|}{97.78} & (97.67, 97.9) & \multicolumn{1}{c|}{98.28} & (98.2, 98.36) \\ \hline
\end{tabular}
\end{center}
\end{table}

\begin{table}[htb]
\caption{Statistical significance evaluation of the different models.}
\label{tab:table_2}
\begin{center}       
\begin{tabular}{|c|c|c|c|}
\hline
Model Comparison & Threshold & Filled Threshold & Hull \\ \hline
Base vs. 1st ref. & \textless 1e-08* & \textless 1e-08* & 1.58e-04*\\ \hline
Base vs. 2nd ref. & \textless 1e-08* & \textless 1e-08* & \textless 1e-08* \\ \hline
Base vs. 3rd ref. & 2.22e-08* & 1.9e-01 & 2.12e-04* \\ \hline
1st ref. vs. 2nd ref. & \textless 1e-08* & \textless 1e-08* & 4.0e-06* \\ \hline
1st ref. vs. 3rd ref. & 4.81e-07* & 1.00e-07* & 2.42e-06* \\ \hline
2nd ref. vs. 3rd ref. & \textless 1e-08* & \textless 1e-08* & \textless 1e-08* \\ \hline
\end{tabular}
\end{center}
\end{table}

\subsubsection{Evaluation on the bad image test set}
We evaluated the deep learning models on the images where the automated segmentation failed to produce acceptable results. For this, the “Bad Mask Testset” set was used; it consists of 33 images that were classified as bad in the second step of the refinement method (Fig.~\ref{fig:fig_3}). Again, ten frames of each sequence are randomly selected, creating a bad image test set of 330 images. Then the base model and refined models are evaluated against it.

The overall performance is good, considering that the dataset only contains images that were initially segmented poorly. The base models achieve an average accuracy of 93.16\% for the threshold model, 96.32\% for the filled threshold model, and 97.93\% for the hull model (Table~\ref{tab:table_3}). All refinement models, besides the third hull refinement, show an even higher accuracy. The 2nd fold refinement models achieve the highest average accuracy of 97.56\% for the threshold model, 97.78\% for the filled threshold model, and 98.13\% for the hull model. When evaluating the statistical significance, all improvements over the base model are significant for the threshold and filled threshold models. For the hull model, the improvement of the second refinement is statistically significant under the Holm-Bonferroni correction (Table~\ref{tab:table_4}). Again it can be seen that the difference between the initial computer vision masking algorithms is largely reduced when comparing the refinement models. The initial accuracy difference of 4.77\% between the threshold and hull base models is reduced to below 1\% for all folds and only 0.57\% for the second refinement.

\begin{table}[htb]
\caption{Average model accuracies (n=6) and 95\% confidence intervals on the "Bad Mask Testset"}
\label{tab:table_3}
\begin{center}       
\begin{tabular}{|c|cc|cc|cc|}
\hline
\multicolumn{1}{|c|}{} & \multicolumn{2}{c|}{Threshold} & \multicolumn{2}{c|}{Filled Threshold} & \multicolumn{2}{c|}{Hull} \\ \cline{2-7} 
\multicolumn{1}{|c|}{Model} & \multicolumn{1}{c|}{mean acc} & \multicolumn{1}{c|}{95\% confid.} & \multicolumn{1}{c|}{mean acc} & \multicolumn{1}{c|}{95\% confid.} & \multicolumn{1}{c|}{mean acc} & \multicolumn{1}{c|}{95\% confid.} \\ \hline
Base & \multicolumn{1}{c|}{93.16} & (92.69, 93.63) & \multicolumn{1}{c|}{96.32} & (96.2, 96.45) & \multicolumn{1}{c|}{97.93} & (97.8, 98.06) \\ \hline
1st ref.& \multicolumn{1}{c|}{97.18} & (97.11, 97.25) & \multicolumn{1}{c|}{97.51} & (97.45, 97.58) & \multicolumn{1}{c|}{98.02} & (97.94, 98.1) \\ \hline
2nd ref.& \multicolumn{1}{c|}{97.56} & (97.48, 97.64) & \multicolumn{1}{c|}{97.78} & (97.7, 97.86) & \multicolumn{1}{c|}{98.13} & (98.06, 98.2) \\ \hline
3rd ref.& \multicolumn{1}{c|}{96.91} & (96.75, 97.08) & \multicolumn{1}{c|}{97.23} & (97.12, 97.35) & \multicolumn{1}{c|}{97.8} & (97.69, 97.92) \\ \hline
\end{tabular}
\end{center}
\end{table}

\begin{table}[htb]
\caption{Statistical significance evaluation of the different models.}
\label{tab:table_4}
\begin{center}       
\begin{tabular}{|c|c|c|c|}
\hline
Model Comparison & Threshold & Filled Threshold & Hull \\ \hline
Base vs. 1st ref. & \textless 1e-08* & \textless 1e-08* & 1.53e-01\\ \hline
Base vs. 2nd ref. & \textless 1e-08* & \textless 1e-08* & 5.06e-03* \\ \hline
Base vs. 3rd ref. & \textless 1e-08* & 6.95e-08* & 9.21e-02 \\ \hline
1st ref. vs. 2nd ref. & 4.42e-06* & 5.94e-05* & 4.0e-06* \\ \hline
1st ref. vs. 3rd ref. & 2.9e-03* & 2.70e-04* & 1.95e-02 \\ \hline
2nd ref. vs. 3rd ref. & 3.48e-06* & 1.37e-06* & 8.08e-05* \\ \hline
\end{tabular}
\end{center}
\end{table}

\subsubsection{Visual evaluation for image de-identification}
In order to assess the ability of the models to remove text, fiducials, and other background data from the input images, the model-generated segmentation masks were also visually assessed. An image that still contained visible text or potential non-ultrasound PHI should be counted as a failed image de-identification. An image that only contains ultrasound data and shows no visible text or potential PHI after the segmentation, independent of the quality of the segmentation itself, should be counted as a successful image de-identification.

\begin{figure}
\begin{center}
\begin{tabular}{c}
\includegraphics[width=0.95\textwidth]{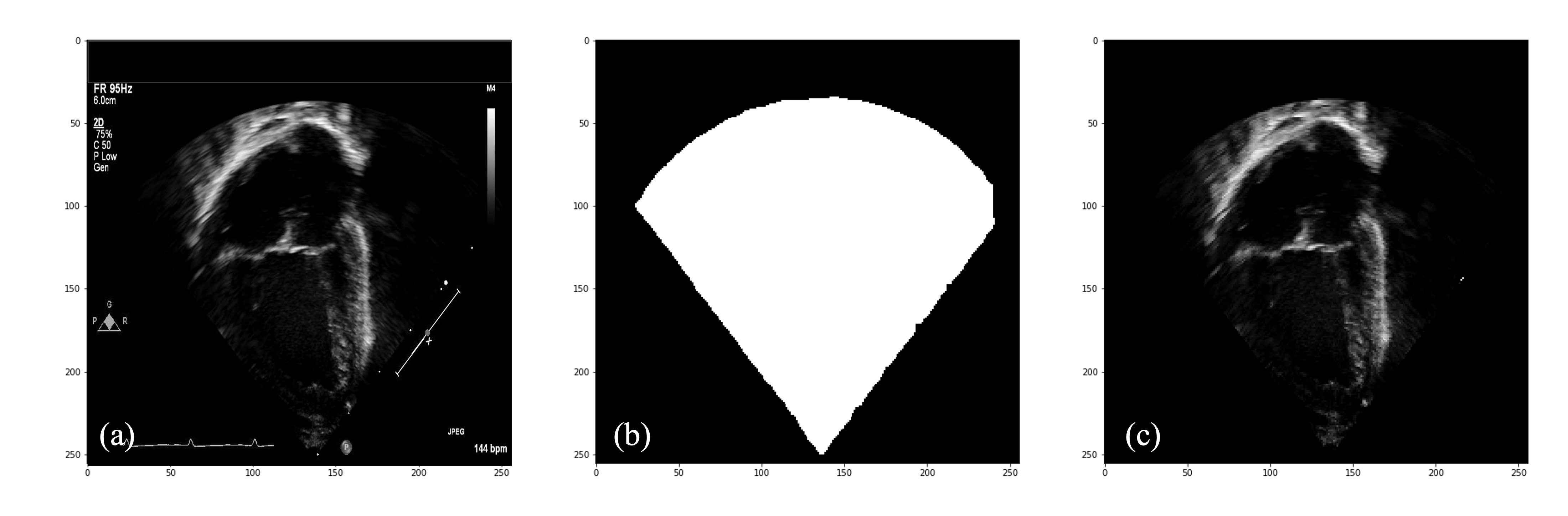}
\end{tabular}
\end{center}
\caption 
{ \label{fig:fig_7}
Example of a successful de-identification (2nd refinement): (a): raw cardiac input frame including multiple burned-in text locations, (b): model generated segmentation mask (c): de-identified masked image.} 
\end{figure}

\begin{figure}
\begin{center}
\begin{tabular}{c}
\includegraphics[width=0.95\textwidth]{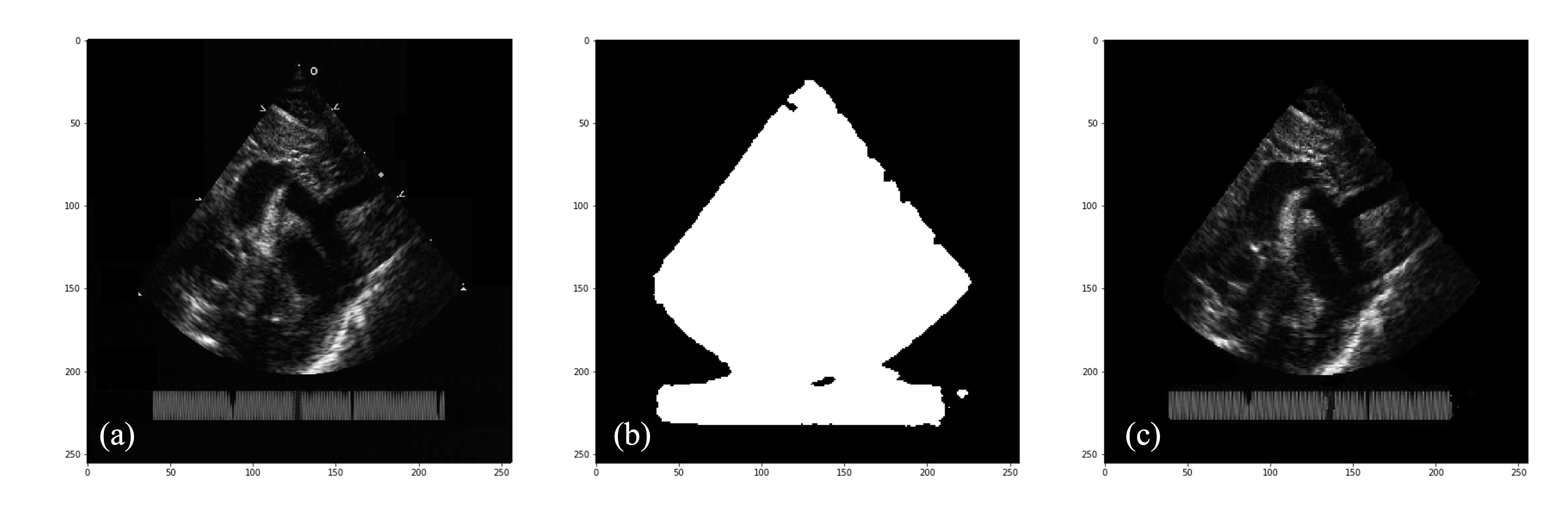}
\end{tabular}
\end{center}
\caption 
{ \label{fig:fig_8}
Example of a failed de-identification (2nd refinement): (a): raw cardiac input frame including a cluttered moving EKG line (b): model generated segmentation mask (c): masked image including the EKG line.} 
\end{figure}

In the visual background removal assessment on the "Bad Mask Testset", all models (Base model, 1st, 2nd, and 3rd refinement) performed equally well. The successfully de-identified 320 frames from burned-in text and potential PHI data. An example of a successful de-identification can be seen in Fig.~\ref{fig:fig_7}. All models also failed on the same ultrasound sequence and all of its' 10 frames. A sample frame of this sequence can be seen in Fig.~\ref{fig:fig_8}. The models, therefore, achieve a de-identification accuracy of 96.97\%, segmenting 320 frames out of 330 correctly. The observed pixel errors all came from dynamic components like EKG. They are therefore unlikely to be PHI, but could still be a potential source of bias when training deep learning models.

\section{Discussion}
\begin{figure}
\begin{center}
\begin{tabular}{c}
\includegraphics[width=0.95\textwidth]{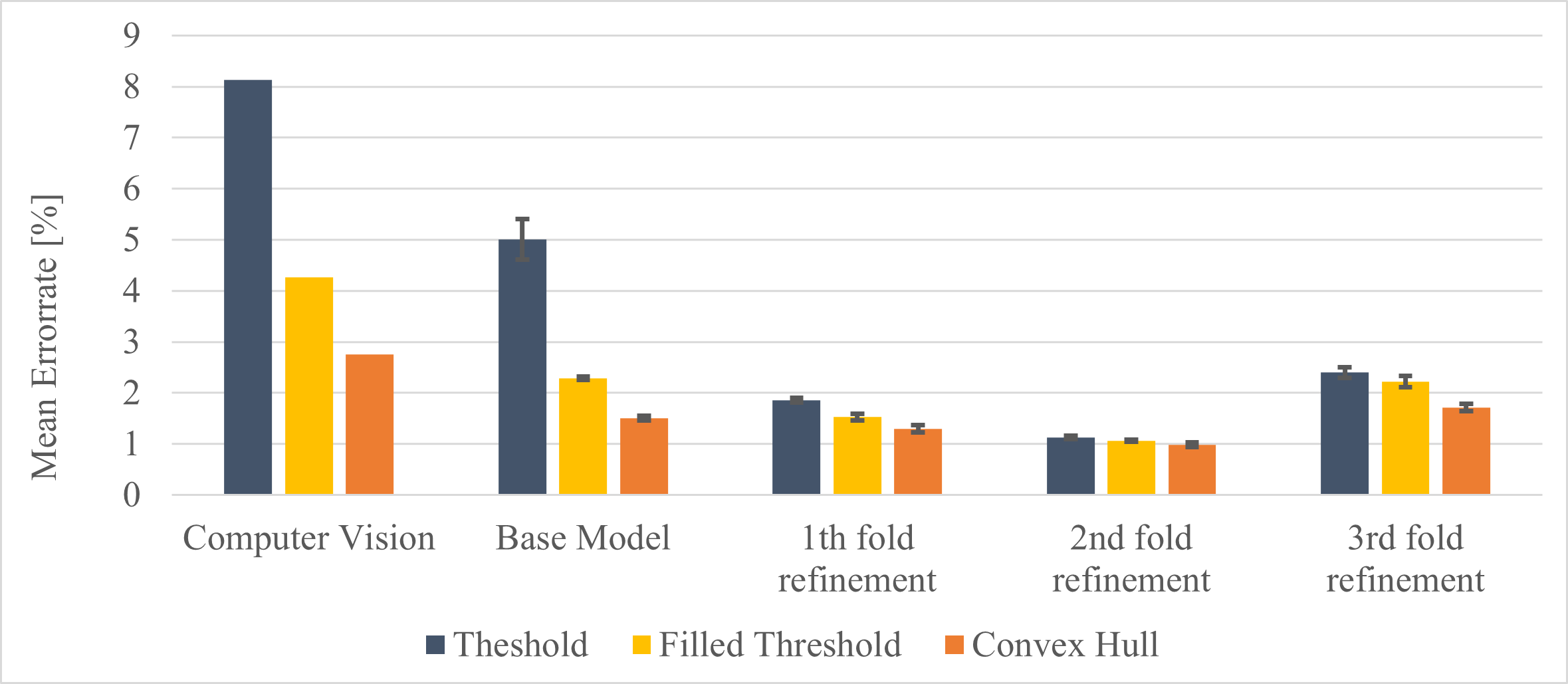}
\end{tabular}
\end{center}
\caption 
{ \label{fig:fig_9}
Comparison of the mean error rates for the computer vision algorithms, base models, and refinement models on the three mask types.} 
\end{figure}

The four-step model refinement method presented in this paper demonstrates a simple and cost-efficient way of increasing segmentation model accuracies. In addition to working well on good quality images, it also improved performance and achieve good accuracies on images that were classified as failures for the computer vision algorithms in the visual check. All refinement model folds, besides the third hull fold, showed a better mean performance, than the base models. This shows that the method is most effective when the user is limited to low accuracy input training masks. With better initial computer vision algorithm performance the method will experience diminishing returns. The presented concept of running deep learning experiments multiple times in order to apply statistical calculations for better experiment evaluation proved to be a valuable tool. In the shown use-case all performance differences, besides the third fold of the filled threshold, are statistically significant. A reason why this fold is significantly different from the others could be that due to the small number of only 22 images used for the refinement, a bad sample would have a substantial negative impact on the performance. More similar results across the folds could probably be achieved by increasing the number of samples.
To show the effectiveness of the proposed refinement process to reduce the necessity to develop complex computer vision algorithms three masking types were compared. The mean error rates (Fig.~\ref{fig:fig_9}) show that the performance of the refined models is much more similar than before. The difference between the refined threshold model and the refined hull model is reduced to only 0.15\% for the second split from an initial 5.37\% in the computer vision performance. This shows that even simple initial training masks, like the threshold mask, can achieve similar results to a more complex input like the hull masks when using model refinement.

There is again a difference in performance between the different refinement folds since the same number of 22 refinement images was used. Therefore, for all models, a bad label in one of the folds would have a strong negative effect on the performance. This would also align with the overall result of the refined models, as the third split performs worst for all mask types. The effect may be reducible by using a larger dataset to refine the base model. But the differences between the refined models in all folds are much smaller than the initial differences of the base models or the computer vision models, therefore proving the huge improvements due to the proposed refinement process.

When evaluating the refined models for their capability to remove text, biasing fiducials, and potential PHI background data in ultrasound images, the method showed excellent results. All models showed great ability to identify the location of the ultrasound cones within the image, only varying in the accuracy of matching the exact shape of the cone. This results in nearly all models being able to separate echo data from the background. All models failed to segment the same testing ultrasound sequence on all ten frames, which included a dynamic EKG. Since the test set is small with only 33 image sequences, this resulted in a noticeable reduction in accuracy for all models. It should therefore be analyzed in future work if the accuracy could be even higher when adding more training samples including dynamic background objects and evaluating the models on a larger test set. The presented de-identification approach is limited though to use-cases where all background data can be removed, and there is no additional value and risk for model bias in keeping non-PHI information burned into the pixel data. For applications where text data needs to be retained in the image, there is still a need for a more targeted approach.

\section{Conclusion}
The results show that the refinement method of training a base model on low-quality automated segmentations and then refining the model on a small number of hand segmented data increases the model performance. By using multiple experimental runs to calculate a p-value, we showed the statistical significance of model improvements. The experiments showed most improvements on rudimentary low-quality input masks and diminishing returns on already well-performing input masks. In our experiments, the differences between more and less sophisticated training masks were reduced by using the refinement method. This shows that this approach reduces the necessity for a large dataset of high-accuracy segmentation masks by utilizing a small accurate dataset only to refine model performance. In our experiments, a difference between the experiment folds remained, but this could be due to the small number of 22 ground truth images that were used to refine the models. We have shown that the utilization of deep learning achieves good performances even in cases where the initial computer vision algorithms fail. The proposed method proved to be a great way to leverage deep learning models. The concept applies to many single cone ultrasound segmentation tasks, where labeling ground truth data is expensive, and human evaluation of automated labels can be fast. For those, the presented experiments also showcase an easy and robust method of removing static text and biasing fiducials, including potential PHI data. The presented approach of running experiments multiple times to allow statistical analysis of the performance improvements, is a useful tool in validating results. If the increased computation costs are not a critical limitation, this enables a more informed way of comparing deep learning experiments.

\section*{Disclosures}
No conflicts of interest, financial or otherwise, are declared by the authors.

\section* {Acknowledgments}
We would like to thank Professor Robert Howe for the helpful discussion and access to computational facilities.


\bibliography{report}   

\begin{thebibliography}{10}

\bibitem{NIPS2012_c399862d}
A.~Krizhevsky, I.~Sutskever, and G.~E. Hinton, ``Imagenet classification with
  deep convolutional neural networks,'' in {\em Advances in Neural Information
  Processing Systems},  F.~Pereira, C.~Burges, L.~Bottou, {\em et~al.}, Eds.,
  {\bf 25}, Curran Associates, Inc.  (2012).

\bibitem{7486599}
S.~Liu and W.~Deng, ``Very deep convolutional neural network based image
  classification using small training sample size,'' in {\em 2015 3rd IAPR
  Asian Conference on Pattern Recognition (ACPR)},  730--734  (2015).

\bibitem{han2016deep}
S.~Han, H.~Mao, and W.~J. Dally, ``Deep compression: Compressing deep neural
  networks with pruning, trained quantization and huffman coding,'' {\em
  International Conference on Learning Representations (ICLR)} {\bf 4}  (2016).

\bibitem{Frankle2018}
J.~Frankle and M.~Carbin, ``The lottery ticket hypothesis: Finding sparse,
  trainable neural networks,'' in {\em 7th International Conference on Learning
  Representations, {ICLR} 2019, New Orleans, LA, USA, May 6-9, 2019},
  OpenReview.net  (2019).

\bibitem{he2016deep}
K.~He, X.~Zhang, S.~Ren, {\em et~al.}, ``Deep residual learning for image
  recognition,'' in {\em Proceedings of the IEEE conference on computer vision
  and pattern recognition},  770--778  (2016).

\bibitem{deng2009imagenet}
J.~Deng, W.~Dong, R.~Socher, {\em et~al.}, ``Imagenet: A large-scale
  hierarchical image database,'' in {\em 2009 IEEE conference on computer
  vision and pattern recognition},  248--255, Ieee  (2009).

\bibitem{Choi2019}
D.~Choi, A.~Passos, C.~J. Shallue, {\em et~al.}, ``Faster neural network
  training with data echoing,'' {\em CoRR} {\bf abs/1907.05550}  (2019).

\bibitem{Harvey2019}
H.~Harvey and B.~Glocker, {\em A Standardised Approach for Preparing Imaging
  Data for Machine Learning Tasks in Radiology: Opportunities, Applications and
  Risks}, 61--72.
\newblock Springer, Cham  (2019).
\newblock [doi:10.1007/978-3-319-94878-2\_6].

\bibitem{Russakovsky2015}
O.~Russakovsky, J.~Deng, H.~Su, {\em et~al.}, ``{ImageNet Large Scale Visual
  Recognition Challenge},'' {\em International Journal of Computer Vision
  (IJCV)} {\bf 115}(3), 211--252  (2015).
\newblock [doi:10.1007/s11263-015-0816-y].

\bibitem{Barth2019}
R.~Barth, J.~IJsselmuiden, J.~Hemming, {\em et~al.}, ``Synthetic bootstrapping
  of convolutional neural networks for semantic plant part segmentation,'' {\em
  Computers and Electronics in Agriculture} {\bf 161}, 291--304  (2019).
\newblock [doi:10.1016/j.compag.2017.11.040].

\bibitem{ronneberger2015}
O.~Ronneberger, P.~Fischer, and T.~Brox, ``U-net: Convolutional networks for
  biomedical image segmentation,''  (2015).
\newblock [doi:10.1007/978-3-319-24574-4\_28].

\bibitem{Kingma2015}
D.~P. Kingma and J.~Ba, ``Adam: {A} method for stochastic optimization,'' in
  {\em 3rd International Conference on Learning Representations, {ICLR} 2015,
  San Diego, CA, USA, May 7-9, 2015, Conference Track Proceedings},  Y.~Bengio
  and Y.~LeCun, Eds.  (2015).

\bibitem{Student1908}
Student, ``The probable error of a mean,'' {\em Biometrika} {\bf 6}(1), 1--25
  (1908).

\end{thebibliography}
\bibliographystyle{spiejour}   


\vspace{2ex}\noindent\textbf{Manuel Zahn} is leading the Machine Learning team at VideaHealth. During this work, he was a researcher at the Harvard Biorobotics Lab. He received his BS and MS degrees in Electrical Engineering and Information Technology from the Technical University of Munich in 2015 and 2019, respectively. He previously conducted research on underwater human-robot interaction at the Stanford Robotics Lab in 2017. His current research interests include robotics, computer vision, and deep learning.

\vspace{2ex}\noindent\textbf{Douglas Perrin} is an Instructor in Surgery at the Harvard Medical School, a Sr. Research Scientist at Boston Children's Hospital, and an Affiliate at the Harvard School of Engineering. He received his BS, MS, and Ph.D. in Computer Science at the University of Minnesota in 1995, 1999, and 2002 respectively. He has published in Computer Vision, Robotics, Visual Servoing, Haptics, ultrasound imaging and procedure guidance, and machine learning for pediatric cardiac pathologies from medical imaging.

\listoffigures
\listoftables

\end{spacing}
\end{document}